# Segmentation and Classification of Cine-MR Images Using Fully Convolutional Networks and Handcrafted Features

M. Hossein Eybposh[1*], Mohammad Haghir Ebrahim-Abadi [2**], Mohammad Jalilpour-Monesi [3**], and Seyed Saman Saboksayr[4**]

Sharif University of Technology, Tehran, Iran
[1]`eybpoosh_mh@ee.sharif.edu` [2]`jalilpourmonesi_m@ee.sharif.edu`
[3]`ebrahimabadi@ee.sharif.edu` [4]`saboksayr_saman@ee.sharif.edu`
*Corresponding Author  **Contributed Equally

**Abstract.** Three-dimensional cine-MRI is of crucial importance for assessing the cardiac function. Features that describe the anatomy and function of cardiac structures (e.g. Left Ventricle (LV), Right Ventricle (RV), and Myocardium(MC)) are known to have significant diagnostic value and can be computed from 3D cine-MR images. However, these features require precise segmentation of cardiac structures. Among the fully automated segmentation methods, Fully Convolutional Networks (FCN) with Skip Connections have shown robustness in medical segmentation problems. In this study, we develop a complete pipeline for classification of subjects with cardiac conditions based on 3D cine-MRI. For the segmentation task, we develop an 2D FCN and introduce Parallel Paths (PP) as a way to exploit the 3D information of the cine-MR image. For the classification task, 125 features were extracted from the segmented structures, describing their anatomy and function. Next, a two-stage pipeline for feature selection using the LASSO method is developed. A subset of 20 features is selected for classification. Each subject is classified using an ensemble of Logistic Regression, Multi-Layer Perceptron, and Support Vector Machine classifiers through majority voting. The Dice Coefficient for segmentation was 0.95$\pm$0.03, 0.89$\pm$0.13, and 0.90$\pm$0.03 for LV, RV, and MC respectively. The 8-fold cross validation accuracy for the classification task was 95.05% and 92.77% based on ground truth and the proposed methods segmentations respectively. The results show that the PPs increase the segmentation accuracy, by exploiting the spatial relations. Moreover, the classification algorithm and the features showed discriminability while keeping the sensitivity to segmentation error as low as possible.

## 1   Introduction

Effective and efficient classification of subjects into different cardiac diseases based on a MR image requires quantitative measures that describe different aspects of cardiac function and anatomy. In a recent systematic review conducted by Kalam et al. [1],

Ejection Fraction (EF) and Global Longitudinal Strain (GLS) [2] were found to have prognostic value in heart failure, acute myocardial infarction, and valvular heart disease. However, these measures are computed from the segmented cardiac structures.

### 1.1 Segmentation

Fully Convolutional Networks (FCN) have shown state of the art performance in many medical image segmentation tasks [3, 4]. A FCN's structure is composed of a contracting path followed by an expanding path. Ronneberger et al [4] introduced the idea of skip connections between the contracting and expanding path, and called the new structure "U-Net" (See Fig. 1).

However, the original U-Net was developed for 2D inputs. Expanding the idea of U-Net to 3D inputs and convolutions suffers from a couple of drawbacks: 1) Using 3D convolutions is more computationally expensive than 2D convolutions. 2) These structures require more GPU memory, since the whole 3D image is loaded on the GPU. 3) More memory means smaller batch size, which makes optimization more difficult. 4) 3D convolution results in a larger number of trainable parameters, which requires larger number of training data to avoid overfitting.

We propose a new structure to segment individual 2D slices while exploiting the 3D relations between adjacent slices. This is done by including skip connections from the contracting path in adjacent slices to the expanding path of the main slice. In this approach, we address all the aforementioned problems by segmenting only 2D slices, and letting the model be spatially aware when making a prediction about each pixel.

### 1.2 Classification

Different feature selection methods and classifiers have been used to classify heart diseases in the literature. McLeod et al [5] have used a simple K-nearest neighbor classifier for classifying Arrhythmogenic Right Ventricular Cardiomyopathy vs. control. Bhatia et al [6] have used integer-coded generic algorithm to select important features and then have used SVM as a classifier. We propose a feature selection pipeline based on L-1 Penalization. We use LASSO in a One vs. Rest structure to cope with the multi-class nature of this classification problem. Finally, we use the ensemble of Support Vector Machine (Nu-SVM), Multi-Layer Perceptron (MLP), and Logistic Regression (LR) for classification.

The article is organized as follows, the segmentation and classification methods are discussed in section 2.1 and 2.2 respectively. The experimental design and results for the segmentation and classification methods are presented in section 3.2 and 3.3. Finally, the discussion about the methods and results are discussed in sections 4 and 5 respectively.

## 2  Methods

### 2.1  Segmentation Using FCN with Parallel Paths

**U-Net.** The basic structure of a U-Net consists of a contracting path and an expanding path. The contracting path consists of a series of Convolutional Blocks (CB), each followed by a Max Pooling layer that reduces the dimensionality of input by a factor of two. The size of the input and output of each CB is outlined by its level L-n, where $n \in \{1, 2, 3, 4, 5\}$ represents the level number. In the expanding path, the same kind of CBs is followed by an un-pooling layer that converts the input size back to its original size. Then the output of the un-pooling layer is concatenated with the output of the CB from the same level in the contracting path (see Fig. 1). The connection from the contracting path to the expanding path is called a Skip Connection (SC).

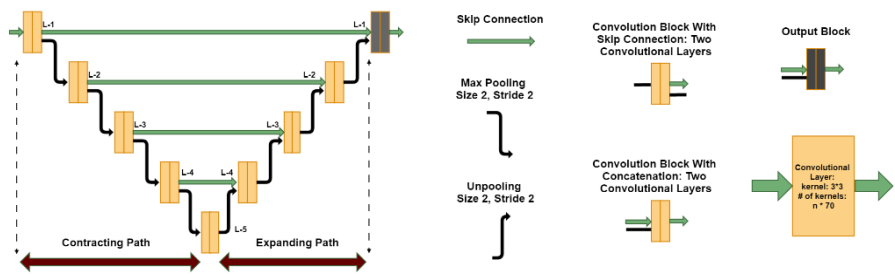

Fig. 1. The structure of a U-Net.

**Connections from Parallel Paths.** Using Parallel Paths (PP), 3D relations are exploited when segmenting individual 2D slices. Fig. 2 shows the proposed structure with PP (called U-Net-PP). In this structure, there are two auxiliary contracting paths, which process one slice before and after the main slice. In other words, when each slice is processed in the main contracting path, the slices that are exactly below and above the main slice are also processed through two separate contracting paths (auxiliary paths). The feature maps of all three contracting paths are concatenated with the features from the expanding path. These connections are called PP connections since they are connecting the feature maps from other spatial locations to the expanding path.

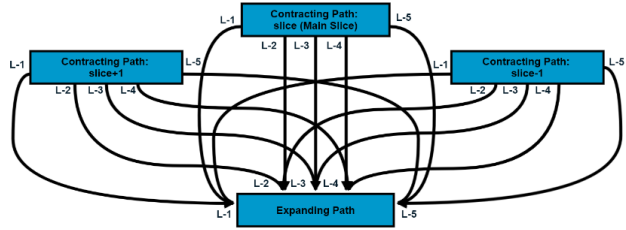

Fig. 2. The structure of the proposed U-Net-PP.

**The Output of the U-Net.** Structures of interest in this study are visually dissimilar, thus using the same convolutional path in the output block for all the structures simultaneously is not the optimum choice. In this study, we will consider four different paths in the output block of the model (i.e. L-1 in the expanding path, shown by gray double-box in Fig. 1). The intuition behind this design is to allow for each path to learn the segmentation rules for BG, LV, MC, and RV individually. The output block that was used in this study is shown in Fig. 3 part a. However, we compare the performance of the different designs.

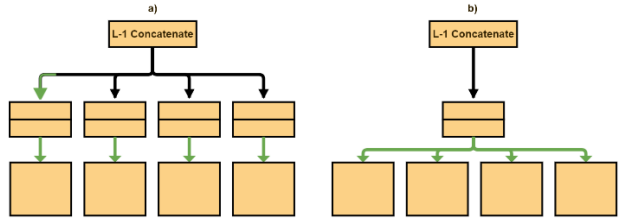

Fig. 3. Different Output blocks. a) the proposed output block b) Simple output block.

**Compensating the Underrepresentation Problem.** Based on initial experiments, we discovered two problems: 1) MC from patients with HCM were harder to segment for U-Net-PP. 2) MC and RV were harder to segment than LV. For MC, this might be due to the underrepresentation of the MC pixels during training. However, for RV, the difficulty might be due to the similarity between LV and RV. To address these problems, we took two approaches: 1) Applying sample weight to subjects with HCM to boost their effect on training the model. 2) Applying pixel weight to pixels that represent the MC and RV to make up for the underrepresentation of the structures, compared to the background.

**Post-Processing the U-Net-PP Output.** In Cine-MRI, there are other organs that look like cardiac structures and might be mistaken by the network. To avoid this confusion in the final predictions, we remove all segmented volumes except for the biggest volume (i.e. heart), from the predicted segmentations. Please note that this problem could also be alleviated by increasing the input size.

**Data Augmentation.** The dataset consists of 100 Three-Dimensional MR images, which consist of 1902 slices in total. In order to avoid overfitting, online data augmentation was applied to the training data. The augmentation included: resizing (randomly zooming in or out), rotation (from -180 to 180 degrees), vertical and horizontal shift and flipping, and shearing.

**Optimization and Training.** The model was implemented in Keras with TensorFlow backend. The ADAM optimizer was used for minimizing the loss function. The negative of Dice Coefficient was used as the loss function. The batch size, learning rate, and number of epochs changed in different stages of training. Batch size ranged from 2 to 20. Number of epochs ranged from 5 to 70, and the learning rate ranged from 1.0e-2 to 1.0e-6. The model was trained on a Nvidia Geforce GTX 1080 graphics card. The training process was finished in 4 hours. Each epoch took less than 120 seconds for a batch size of 2 and 60 seconds for a batch size of 16.

## 2.2 Classification

**Features.** In order to describe the anatomy and function of the heart, 125 hand-crafted features were extracted the segmentations of cardiac structures. The features employed in this study are divided into three groups, Volumetric, Thickness, and Shape features. A brief description of these features is provided in the following:

*Volumetric.* As mentioned in the introduction section, features based on structure volumes are key indicators of cardiac dysfunction. Therefore, the volume of right and left ventricles at the end of systole and diastole and their ratio were considered in the feature pool. Another feature based on volume is ejection fraction of LV and RV.

*Thickness.* Thicknesses on every angle were estimated from the segmentation of MC. Next, we evaluated the maximum, minimum, mean, median, standard deviation, and variance of thicknesses, and the number of thicknesses that were higher than THR millimeters. THR is threshold and ranged from 10 to 30 mm in 1mm steps.

*Shape.* These features include:
Surface area and ratio of surface area to volume
Sphericity: measures the similarity of a structure and a sphere.
Compactness 1,2: like sphericity, is a measure of similarity between a structure and a sphere.
Spherical disproportion: surface area of a structure divided by surface area of a sphere with the same volume.
Maximum 3D diameter: the biggest pairwise Euclidean distance between structure's voxels.
Maximum 2D diameter (for slice, column, and row): the biggest pairwise Euclidean distance between structure's voxels in slice, column and row respectively.
Major, Minor, Least axis:

$$major\ axis = 4\sqrt{\lambda_{major}} \qquad (1)$$

$$minor\ axis = 4\sqrt{\lambda_{minor}} \qquad (2)$$

$$least\ axis = 4\sqrt{\lambda_{least}} \qquad (3)$$

Where $\lambda_{major}$, $\lambda_{minor}$ and $\lambda_{least}$ are the lengths of largest, second largest and smallest principal component axes.
Elongation: is square root of the ratio of $\lambda_{minor}$ to $\lambda_{major}$.
Flatness: square root of the ratio of $\lambda_{least}$ to $\lambda_{major}$ [7].

**Feature selection and classification.** We compare the performance of three different feature selection methods that are based on L-1 penalization of the coefficients, namely LASSO regression, Logistic regression with L-1 penalization, and Randomized Logistic regression. To extend these methods to the multi-class classification problems, we use the One-vs-Rest approach.

The LASSO regression estimates the coefficients by forcing some of them to be exactly equal to 0 and shrinking the others. Consider data $(X^i, Y^i)$, i=1, ..., N where $X^i=(x_{i1}, ..., x_{ip})$ and $Y^i$ are independent and outcome variables respectively

$$\hat{\beta}=\operatorname{argmin}(\ \|Y-\beta X\|_2^2 + \lambda \sum_{j=1}^{p} |\beta_j|\ ) \qquad (4)$$

Using equation (4) regression coefficients $\hat{\beta}=(\hat{\beta}_1, ..., \hat{\beta}_p)$ are estimated [8]. Parameter λ determines the shrinkage level.

Contrary to the linear regression method, that fits linear model to data, the logistic regression method uses nonlinear function to regress the data. The logistic regression model is defined as follows:

$$Prob(b|X) = \frac{1}{1+(-b(w^T X+v))} \qquad (5)$$

Where, $b \in \{1, -1\}$, w is the coefficient vector, v is the intercept, and $Prob(b|X)$ is the conditional probability of b, given $X \in R^p$. The l1-Regularized logistic regression problem is:

$$\text{minimize}\ \ (\tfrac{1}{N}) \sum_{i=1}^{N} f(w^T b_i X^i + v b_i) + \lambda \sum_{j=1}^{p} |w_j| \qquad (6)$$

Where $f(z)=\log(1+\exp(-z))$, and λ is the regularization parameter. Just like LASSO, as the value of λ increases, more elements of coefficient vector become zero [9].

Randomized logistic regression uses logistic regression to select a set of features based on stability path. The stability path depicts the probability of selecting a feature during random resampling from the data. In fact, randomized logistic regression is expected to select a more stable set of features than logistic regression because it considers different possibilities both in random subsampling and method parameters.

The regularization and classifier parameters were determined through a grid search process. For each set of the aforementioned parameters, the mean and standard deviation of the classification accuracy in 8 random 8-fold cross-validation are considered as the parameter selection criteria (i.e. highest mean and lowest variance). In the first step, to select the best set of features, 30 features from the thickness shape feature sets were selected based on the frequency of selection through the process mentioned process.

Next, these features are combined with the 12 volumetric features. Since L-1 penalized methods yield uncorrelated features, we believe that the features in the new set are less correlated than the initial features. In the second stage, the same feature selection procedure is applied on the new feature set and the top 20 features are selected for classification.

We use the ensemble of Nu-SVM, MLP, and LR through soft majority voting for classification. In the voting process, Nu-SVM was weighted twice the other classifiers. Nu-SVM has a sigmoid kernel, and MLP has one hidden layer with 10 nodes. In the next section, we discuss the results of our segmentation and classification methods.

## 3 Results

### 3.1 Segmentation Experiments

To validate the performance of our segmentation method, we performed 5-Fold Cross Validation on subjects. In other words, at each fold, 80 subjects are randomly selected for training the model, and the remaining 20 are used for validation. The 5-fold cross validation results for the proposed method are presented in Table 1.

As seen in Table 1, U-Net-PP increases the segmentation performance of the U-Net based model. When considering the LV, although a decrease in the Dice Coefficient is observed in PP, we also observe a lower variance. Moreover, the U-Net-PP outperforms the simple U-Net in MC and RV segmentation in terms of Dice Coefficient. Considering the Hausdorff Distance, the results indicate that the U-Net-PP follows the contour of the GT segmentations, better than a simple U-Net. However, a statistical hypothesis test to compare the results of the outputs of the two models was not applicable because of the small sample size (only 20 samples in the validation set).

Table 1 Performance metrics of different segmentation methods and structures. LV, MC, and RV stand for Left Ventricle, Myocardium, and Right Ventricle respectively.

|           | Dice Coefficient |             |             | Hausdorff Distance |              |              |
|-----------|-----------------|-------------|-------------|--------------------|--------------|--------------|
| Structure | LV              | MC          | RV          | LV                 | MC           | RV           |
| U-Net-PP  | 0.95±0.03       | **0.90±0.03** | **0.89±0.13** | **5.8±2.1**      | **16.0±12.2** | **20.7±10.6** |
| U-Net     | **0.96±0.08**   | 0.87±0.09   | 0.88±0.06   | 6.0±4.2            | 18.0±12.6    | 25.7±25.0    |

## 3.2 Classification Experiments

Table 2 shows the performance of different feature selection methods based on the result of ensemble classification on U-Net-PP and Ground Truth segmentations. The results show that the LASSO method outperforms the other two. Comparing the results from Table 2 shows that by using the U-Net-PP segmentations result, the classification accuracy drops less than 3%. This indicates that the features, feature selection process, and the classifiers are robust to errors in the segmentation of the structures.

Table 2. Classification accuracy for different feature selection methods on U-Net-PP and GT segmentations.

| Accuracy | LASSO | Log. regression | R. Log. regression |
|---|---|---|---|
| U-Net-PP | **0.93±0.06** | 0.91±0.07 | 0.90±0.08 |
| GT | **0.95±0.05** | 0.92±0.07 | 0.91±0.07 |

## 4 Conclusion and Discussion

In this study, we develop a complete pipeline for the classification of subjects into four groups of cardiac dysfunctions and healthy subjects, based on their Cine-MR images. The pipeline includes a segmentation model based on fully convolutional networks and skip connections. The model exploits the special relation through PP skip connections. The output of the model is then post processed to eliminate the disconnected parts from the result. A number of features are extracted from the segmented volumes and then classified accordingly, through an ensemble of Support Vector Machine (Nu-SVM), Multi-Layer Perceptron (MLP), and Logistic Regression classifiers with soft majority voting. The results of the pipeline are promising, both for segmentation and classification tasks.

The results indicate that PPs are an effective way to exploit the 3D structure of the heart. The classification results show a small drop in the classification accuracy when switching from the ground truth segmentations to the predicted segmentations. Thus, the authors believe that the features are describing different classes in a robust way. However, the authors believe that further investigation is required to validate the conclusion about the effectiveness of PPs in 3D segmentations. For example, the same segmentation approach can be applied to other segmentation problems such as the segmentation of brain or lung tumors. Moreover, further studies are needed to investigate how each disease affects the features, and what aspects of each disease the features are describing.

## References

1. Kalam, K., Otahal, P., Marwick, T.H.: Prognostic implications of global LV dysfunction: a systematic review and meta-analysis of global longitudinal strain and ejection fraction. Heart 100, 1673-1680 (2014)

2. Reisner, S.A., Lysyansky, P., Agmon, Y., Mutlak, D., Lessick, J., Friedman, Z.: Global longitudinal strain: a novel index of left ventricular systolic function. Journal of the American Society of Echocardiography 17, 630-633 (2004)
3. Milletari, F., Navab, N., Ahmadi, S.-A.: V-net: Fully convolutional neural networks for volumetric medical image segmentation. In: 3D Vision (3DV), 2016 Fourth International Conference on, pp. 565-571. IEEE, (Year)
4. Ronneberger, O., Fischer, P., Brox, T.: U-net: Convolutional networks for biomedical image segmentation. In: International Conference on Medical Image Computing and Computer-Assisted Intervention, pp. 234-241. Springer, (Year)
5. McLeod, K., Wall, S., Leren, I.S., Saberniak, J., Haugaa, K.H.: Ventricular structure in ARVC: going beyond volumes as a measure of risk. Journal of Cardiovascular Magnetic Resonance 18, 73 (2016)
6. Bhatia, S., Prakash, P., Pillai, G.N.: SVM based decision support system for heart disease classification with integer-coded genetic algorithm to select critical features. In: Proceedings of the world congress on engineering and computer science, pp. 34-38. (Year)
7. Aerts, H.J.W.L., Velazquez, E.R., Leijenaar, R.T.H., Parmar, C., Grossmann, P., Cavalho, S., Bussink, J., Monshouwer, R., Haibe-Kains, B., Rietveld, D.: Decoding tumour phenotype by noninvasive imaging using a quantitative radiomics approach. Nature communications 5, (2014)
8. Tibshirani, R.: Regression shrinkage and selection via the lasso. Journal of the Royal Statistical Society. Series B (Methodological) 267-288 (1996)
9. Koh, K., Kim, S.-J., Boyd, S.: An interior-point method for large-scale l1-regularized logistic regression. Journal of Machine learning research 8, 1519-1555 (2007)